%% file: main.tex
\documentclass[letterpaper]{article} 
\usepackage{aaai2026}  
\usepackage{times}  
\usepackage{helvet}  
\usepackage{courier}  
\usepackage[hyphens]{url}  
\usepackage{graphicx} 
\usepackage{natbib}  
\usepackage{caption} 
\usepackage{booktabs}
\usepackage{multirow}
\usepackage{amsmath}
\usepackage{amssymb}
\nocopyright
\usepackage{algorithm}
\usepackage{algorithmic}
\usepackage{hyperref}
\usepackage{amsthm}

\definecolor{lightblue}{RGB}{27, 63, 181}
\definecolor{lightred}{RGB}{219, 15, 121}
\hypersetup{
    colorlinks=true,
    linkcolor=lightblue,   
    urlcolor=lightblue,    
    citecolor=lightblue,   
    pdfborder={0 0 0} 
}

\usepackage{listings}
\DeclareCaptionStyle{ruled}{labelfont=normalfont,labelsep=colon,strut=off} 
\floatstyle{ruled}
\newfloat{listing}{tb}{lst}{}
\floatname{listing}{Listing}

\newtheorem{theorem}{Theorem}
\newcommand{\model}{SAGMM}

\title{Self-Adaptive Graph Mixture of Models}

\author{
  Mohit Meena,
  Yash Punjabi,
  Abhishek A,
  Vishal Sharma\textsuperscript{1},
  Mahesh Chandran\\[6pt]
  {\normalfont \small \
     Fujitsu Research of India, Bangalore\\
    \{mohitkumar.meena, yash.punjabi, abhishek.a, mahesh.chandran\}@fujitsu.com,
    i.vishal1990@gmail.com
  }
}

\begin{document}
\maketitle
\footnotetext[1]{ Work done while at Fujitsu Research of India, Bangalore. Now at Microsoft.}
\begin{abstract}
Graph Neural Networks (GNNs) have emerged as powerful tools for learning over graph-structured data, yet recent studies have shown that their performance gains are beginning to plateau. In many cases, well-established models such as GCN and GAT, when appropriately tuned, can match or even exceed the performance of more complex, state-of-the-art architectures. This trend highlights a key limitation in the current landscape: the difficulty of selecting the most suitable model for a given graph task or dataset. To address this, we propose Self-Adaptive Graph Mixture of Models (\model{}), a modular and practical framework that learns to automatically select and combine the most appropriate GNN models from a diverse pool of architectures. Unlike prior mixture-of-experts approaches that rely on variations of a single base model, \model{} leverages architectural diversity and a topology-aware attention gating mechanism to adaptively assign experts to each node based on the structure of the input graph. To improve efficiency, \model{} includes a pruning mechanism that reduces the number of active experts during training and inference without compromising performance. We also explore a training-efficient variant in which expert models are pretrained and frozen, and only the gating and task-specific layers are trained. We evaluate \model{} on 16 benchmark datasets covering node classification, graph classification, regression, and link prediction tasks, and demonstrate that it consistently outperforms or matches leading GNN baselines and prior mixture-based methods, offering a robust and adaptive solution for real-world graph learning. Code is released at {\hypersetup{urlcolor=lightred}%
\href{https://github.com/ast-fri/SAGMM}{https://github.com/ast-fri/SAGMM}}.
\end{abstract}
\section{Introduction}

Graph Neural Network (GNN) architectures have seen substantial progress since their early formulation in seminal work on graph-based semi-supervised learning \cite{kipf2016semi}. Though developments in the neural architecture have led to an improved performance (on an average), there is still no consensus on distinct advantage of one model over others. This is reflected in two important observations in recent times: (a) the performance of GNN models have plateaued with no significant performance gain reported with complex architecture. (b) slight tuning of hyperparameters of classical GNN models like GCN \cite{kipf2016semi}, GAT \cite{velivckovic2017graph}, and GraphSAGE \cite{hamilton2017inductive} have shown state-of-the-art (SOTA) performance in node classification tasks, matching or even surpassing latest Graph Transformers (GTs) across diverse datasets \cite{luo2024classic}. This suggests different models tend to learn different (overlapping) regions of the representation space, but each falling short of covering the space to capture diverse patterns and features of the datasets.
Moreover, selecting the right model for a dataset often involves trial-and-error and computationally expensive hyperparameter tuning, with many models discarded after underperforming. However, each model may still capture unique structural patterns or inductive biases (model assumptions) that, while insufficient alone, could contribute meaningfully when combined.
In machine learning (ML), ensemble learning approach is a proven method to pool a set of (weak) models to create a strong model \cite{li2021neural}. In recent years, mixture of experts (MoE) approach has become the \textit{de facto} choice for ensemble learning, though it differs from the classical approach by the gating (routing) network which is trained to select expert(s) based on the input context. The MoE and its variants have been successfully applied to large language models (LLMs) to increase model capacity without proportionate increase in computational cost \cite{fedus2022switch}.
In the graph domain, MoE-based approaches are still in the early stages of exploration but have demonstrated significant potential. Existing methods such as GMoE~\cite{wang2023graphmixtureexpertslearning} typically exhibit limited architectural diversity, often employing variations of a single base model (e.g., GCNs with MoE layers inserted at intermediate depths). Furthermore, their gating mechanisms are generally not designed to capture complex graph topologies, and expert pruning strategies, if present, are either ad hoc or inefficiently integrated. These limitations underscore the need for a more flexible and topology-aware MoE framework tailored to the unique challenges of graph-structured data.
\\
In this work, we propose Self-Adaptive Graph Mixture of Models (\model{}), a modular and practical framework that learns to automatically select and combine the most appropriate GNN models for each part of the graph. \model{} leverages a diverse pool of GNN architectures, each serving as an expert with a fixed view of the data, and employs a topology-aware sparse attention gate that dynamically routes input to the most relevant experts. The framework is designed to be trained end-to-end but as a practical variation, we also explore a training strategy, where experts are pre-trained independently and only the routing mechanism and task head are trained, reducing training time.
The core idea of \model{} is that different GNNs offer complementary perspectives of the graph, but their architectural rigidity limits their individual expressiveness. By selectively combining them at the representation level, \model{} achieves greater adaptability and structural awareness. This formulation also aligns with the broader model merging paradigm \cite{yang2024model}, though our merging occurs in representation space rather than parameter space. The salient features of the \model{} are as follows:
\begin{enumerate}
    \item \model{}  leverages the strength of diverse models such as GCN, GAT, and GraphSAGE, \textit{etc.} each contributing unique inductive biases to capture varied structural properties.
    \item \model{} incorporates a sparse, topology-aware attention gating that adaptively selects the most relevant experts for each node by evaluating attention scores between the node and the expert models.
    \item Pruning of expert pool using importance score which improves efficiency while preserving performance.
    \item Pooling pre-trained models that are otherwise discarded during the model selection phase of standard real-world pipelines unlocks additional value within the SAGMM architecture and enables flexible integration of newer models at minimal cost.
\end{enumerate}
\noindent
Extensive experiments on node classification, graph classification, graph regression, and link prediction tasks demonstrate that \model{} consistently outperforms GNN models and prior graph-based MoE frameworks. Additionally, we also provide theoretical analysis of the framework.

\section{Background and Related Work}
\subsection{Preliminaries and Notations}
Let \( G(A, X) \) be an undirected graph with an adjacency matrix \( A \in \{0,1\}^{n \times n} \) and a node feature matrix \( X \in \mathbb{R}^{n \times d} \), where \( n \) is the number of nodes and \( d \) is the feature dimension. \(D\) represents the degree matrix of graph. The batch size is denoted as \( b \). In a GNN with \( L \) layers, let \( C^{(l)} \) denote the convolution matrix and \( W^{(l)} \) the learnable parameters at layer \( l \).\\
We define a pool of experts \( e = \{e_1, e_2, \dots, e_{N_0}\} \), where \( N_0 \) is the initial number of experts and \( N \) is the number of experts after pruning, such that \( N = N_0 \) at the start of training. For a given node \( u \), let \( k_u \) denote the subset of experts it activates. Notably, in conventional models, the expert pool remains fixed (\( N = N_0 \)).

\label{related_work}
\subsection{Graph Neural Networks}

Several GNN architectures based on encoder-decoder framework \cite{hamilton2017representation} with encoding performed by message-passing mechanism have been proposed for graph ML tasks. According to \cite{balcilar2021analyzing}, the general update rule for the node embeddings 
$H^{(l+1)}$ at layer $l+1$ is represented by:
\begin{equation}
H^{(l+1)} = \sigma \left( \sum_q C^{(l,q)} H^{(l)} W^{(l,q)} \right)   
\label{forwardprop}
\end{equation}
where $\sigma(\cdot)$ is a nonlinear activation function, $W^{(l,q)}$ are learnable parameters associated with the filter $q$, and $C^{(l,q)}$ are model-specific convolution matrices. 
Different GNNs define $C^{(l,q)}$ differently. For instance,  GCN~\cite{kipf2016semi} uses a normalized adjacency matrix fixed across layers, GraphSAGE~\cite{hamilton2017inductive} uses identity and mean aggregation, GAT~\cite{velivckovic2017graph} introduces attention-based weighting, while GIN~\cite{xu2018powerful} approximates the Weisfeiler-Lehman (WL) test via learnable aggregation~\cite{ding2022sketchgnn}. JKNet~\cite{xu2018representation} improves depth flexibility by aggregating information from multiple GNN layers via jump connections.
Variants such as SGC~\cite{wu2019simplifyinggraphconvolutionalnetworks} eliminate nonlinearities for efficiency, MixHop~\cite{abu2019mixhop} aggregates multi-hop neighborhoods, and GraphCNN~\cite{defferrard2016convolutional} applies spectral filtering via Chebyshev polynomials.

\subsection{Mixture of Experts in Graph Learning}
\label{moe_graph_ref}
The MoE framework \cite{jacobs1991adaptive}, when first introduced enables routing inputs to specialized models via a trainable dense gating network. Modern sparse activation MoE architectures, widely used in large-scale language and vision models, leverage thousands of experts to enhance model capacity \cite{zhang2025mixture}. Methods such as DynMoE \cite{guo2024dynamic} introduces top-any gating and pruning strategy to dynamically activate and prunes expert in LLM domain. While M32 \cite{wu2024yuan} introduces an attention-based routing mechanism, its quadratic complexity limits scalability for large graphs. Recent surveys~\cite{cai2025survey, zhang2025mixture} provide a comprehensive overview of trends and challenges in MoE models for LLMs.
\\
For graph data, GMoE \cite{wang2023graphmixtureexpertslearning} extends MoE to GNNs, enabling multi-hop information aggregation. GraphMETRO \cite{wu2024graphmetro} leverages MoE to tackle distribution shifts in GNNs. TopExpert \cite{kim2023learningtopologyspecificexpertsmolecular} employs domain-specific linear experts for graph classification but lacks adaptability to diverse tasks. DA-MoE \cite{yao2024moe} uses GNN layers as experts to address depth sensitivity but does not support dynamic expert selection. GraphMoRE \cite{guo2025graphmore} leverages a Mixture of Riemannian Experts, where each expert operates on a learned manifold to capture diverse local structures. Link-MoE \cite{ma2024mixture} relies on pre-trained GNN experts for link prediction, limiting its flexibility for broader applications.

\section{Methodology}
\label{methodology}
\begin{figure*}[!htbp]
  \centering
  \includegraphics[width=0.87\textwidth]{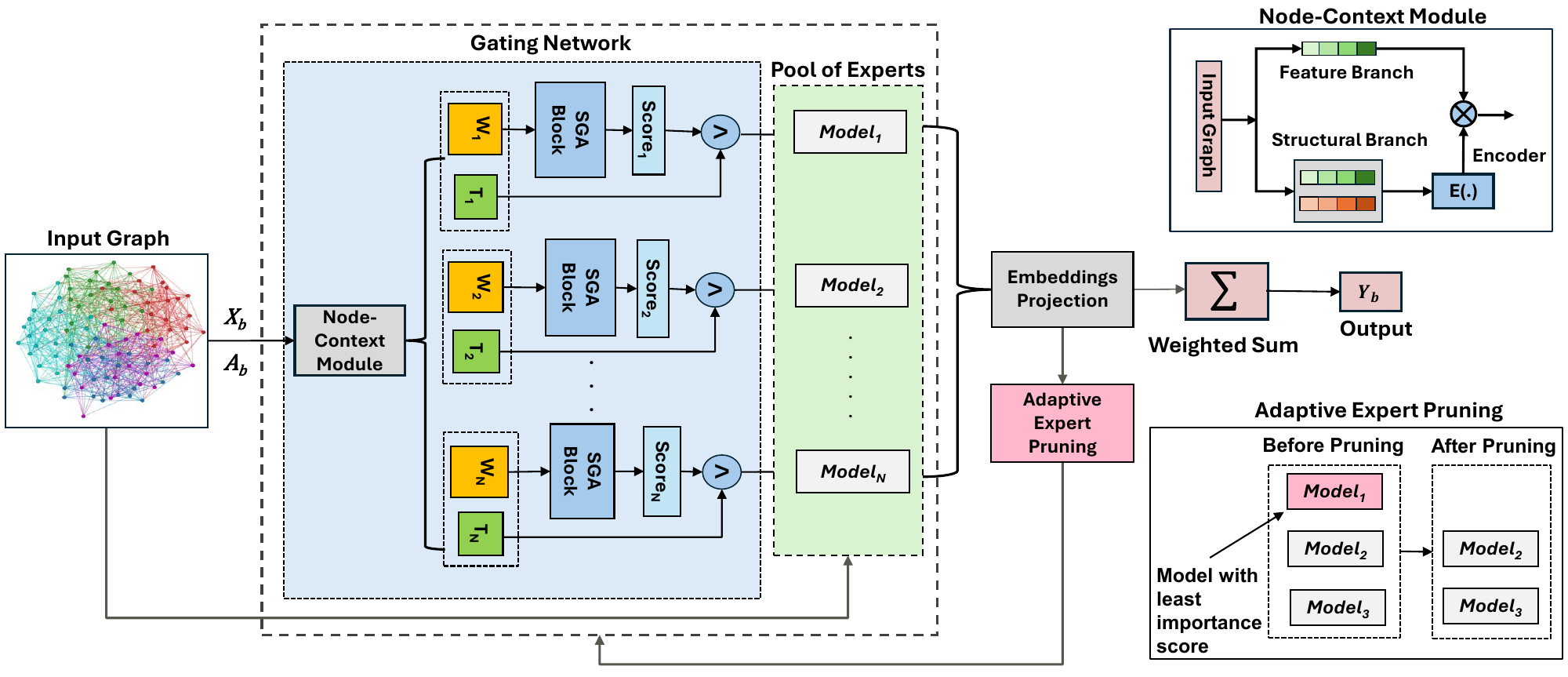} 

  \caption{The overall illustration of the \model\ framework. The key components includes the gating network, a pool of experts, and an adaptive expert pruning. $W_i$ denotes $W_{Q_{i}}$, $W_{K_{i}}$, and $W_{V_{i}}$ per expert $i$.}
  \label{fig:architecture}
\end{figure*}

\label{section_diverse_model}

We now introduce the \model{} in detail, with the subsections focusing its key components.
Figure~\ref{fig:architecture} illustrates the \model{} architecture. Input features are first enhanced with structural context, and expert selection is performed by a gating network using Simple Global Attention (SGA) \cite{wu2023sgformer} and learnable thresholds. The selected expert outputs are projected into a shared space, weighted by gate scores, and aggregated, while low-importance experts are pruned based on an adaptive threshold $\eta$.
The training phase of SAGMM in outlined in Algorithm \ref{alg:SAGMM}.

\subsection{Pool of Experts}

A distinctive feature of the \model{} framework lies in its use of a structurally diverse set of expert models. This choice is motivated by the aim of maximizing architectural heterogeneity, thereby enhancing the expressiveness and representational capacity of the framework. \\
Recent surveys~\cite{chen2020bridging, ju2024comprehensive} categorize GNNs along three key dimensions: propagation strategy (spectral vs.\ spatial), aggregation mechanism (mean vs.\ attention), and training setup (transductive vs.\ inductive). Guided by these dimensions, we include GNNs like GCN~\cite{kipf2016semi} for its spectral filtering and suitability for homophilic graphs, GAT~\cite{velivckovic2017graph} for attention-based neighbor weighting, and GraphSAGE~\cite{hamilton2017inductive} for its inductive capability through neighborhood sampling, along with additional models in the expert pool. Such heterogeneity enables the router to select experts that align with the structural and statistical properties of individual instances.

Formally, the embedding update for an expert $e_i$ at layer $l+1$ is given by:
\begin{equation}
    H_{e_i}^{(l+1)} = f_{e_i}(H_{e_i}^{(l)}, A),
\end{equation}
Here $f_{e_i}$ denotes the message-passing function of expert $e_i$.

This design is also motivated by the bias-variance tradeoff \cite{yang2020rethinking}, where over-parameterized models can improve generalization by reducing variance. By incorporating multiple GNNs, our framework benefits from this property. Moreover, in line with the No Free Lunch theorem \cite{goldblum2023no} which states that no single model performs optimally across all tasks, our approach accounts for variations in network topology, homophily, and heterophily that affects GNN performance.
We show empirically that by pooling experts, each with an unique message-passing mechanism, within MoE framework, our approach dynamically adapts to diverse graph structures, overcoming limitations of a single model.

\subsection{Gating Network}
\label{section_gating}
\begin{figure}[!t]
    \centering
    \includegraphics[width=\linewidth]{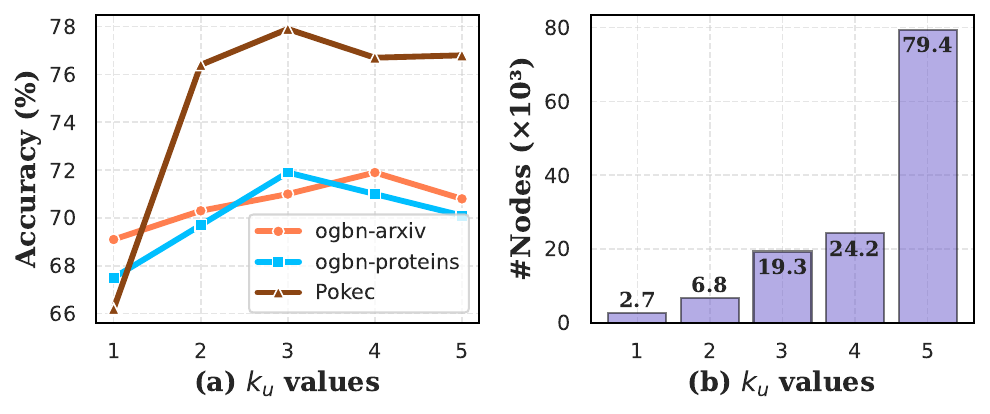}
    \caption{(a) Performance variation across different Top-$k$ values in GMoE-GCN for various datasets.  (b) Distribution of expert activation counts by \model{} for ogbn-proteins dataset, which contains 132{,}534 nodes.
}
    \label{fig:side_by_side_figures}
\end{figure}

In this subsection, we introduce \textit{Topology-Aware Attention Gating (TAAG)}, a novel gating mechanism for graph data.  We begin by highlighting the limitations of existing methods before presenting our proposed approach. \\
\textbf{Limitations of traditional gating mechanism.} 
Several approaches have been proposed for designing gating networks in MoE architectures~\cite{jordan1994hierarchical, clark2022unified, zhou2202mixture}. A widely adopted sparse gating strategy is the noisy top-\(k\) gating mechanism~\cite{shazeer2017outrageously}, which applies a softmax over the top-\(k\) logits produced by a linear transformation of the input. To encourage expert diversity and prevent load imbalance, this approach injects learnable Gaussian noise into the logits, a modification later adapted for graph data in GMoE model.
\\
Despite considerable success of the noisy top-$k$ gating in improving training and inference efficiency, our analysis reveals several key limitations still remain:

\begin{enumerate}
    \item The model’s performance is highly dependent on the choice of \( k \), as different datasets exhibit varying optimal values as shown in Figure~\ref{fig:side_by_side_figures}(a) for GMoE-GCN \cite{wang2023graphmixtureexpertslearning}, highlighting the need for extensive hyperparameter tuning. Moreover, using a fixed \( k \) forces all nodes to activate the same number of experts, which may not be needed considering structural and feature characteristics. 
    As shown in Figure~\ref{fig:side_by_side_figures}(b), different nodes activate different number of experts as per their requirement, eventually leading to performance gains.
  
    \item While prior works on gating strategies select multiple experts, thus implicitly including correlation between them, it is essential to ensure diversity among them for better coverage of the representation space.

     \item Existing methods often rely on simple linear projections of node features to compute gating scores, neglecting structural dependencies within the graph. While DA-MoE \cite{yao2024moe} and GraphMoRE \cite{guo2025graphmore} attempt to overcome this limitation by leveraging GNN-based gating networks, a fundamental drawback remains: these approaches are still unable to effectively capture global graph structural information, which is vital for informed expert selection.
\end{enumerate}

To address these limitations, for each node, TAAG selects experts based on
the local and global structural information while dynamically adjusting the number of activated experts to improve efficiency and robustness. We leverage the SGA block from SGFormer \cite{wu2023sgformer} to compute attention scores efficiently. While standard attention mechanisms have a quadratic complexity of \(O(n^2)\), SGA operates in linear time \(O(n)\) w.r.t. $n$ nodes, making it scalable for medium to large-scale graphs. To incorporate global structural information, we construct the matrix \( X^{(g)} \) using the \( p \) smallest eigenvectors of the normalized graph Laplacian \( L \) (Eq.~(\ref{lap_eqn})), providing positional encodings for each node~\cite{rampasek2022GPS}. We do not use diffusion-based methods \cite{gasteiger2019diffusion, chamberlain2021grand}
as they involve expensive preprocessing, assumes homophily and perform poorly in tasks like link prediction. To further strengthen the structural information, we also add local structural features by aggregating 1-hop and 2-hop (approximate) features \cite{hamilton2017inductive, wang2020multi}. The detailed formulations are provided in Eq.~(\ref{multihop_eq}) and Eq.~(\ref{lap_eqn}).

\begin{equation}
\label{multihop_eq}
    \mathbf{X}^{(1)} = D^{-1} A \mathbf{X}, \quad \mathbf{X}^{(2)} = (D^{-1} A)^2\mathbf{X}.
\end{equation}  
\begin{equation}
\label{lap_eqn}
    L = I - D^{-1/2} A D^{-1/2}.
\end{equation}
Hence, we define the final input features as:  
\begin{equation}
\label{eq: toplogical_aware}
    \mathbf{X'} = \underbrace{\frac{1}{3} (\mathbf{X} + \mathbf{X}^{(1)} + \mathbf{X}^{(2)})}_{\text{Local Term}} \quad || \underbrace{\mathbf{X}^{(g)}}_{\text{Global Positioning of Nodes}}.
\end{equation}  
where $X' \in \mathbb{R}^{n \times (d+p)}$. The local term aggregates multi-hop neighborhood information, while
$X^{(g)}$ denotes the global positional encodings of the nodes of the graph. 
To compute gating scores, we first define the Query (Q), Key (K), and Value (V) representations as follows:
\begin{equation}
    Q = X' W_Q, \quad K = X'W_K, \quad V = X'W_V
\end{equation}
where:  \( W_Q, W_K, W_V \in \mathbb{R}^{(d+p) \times N} \) are learnable projection matrices for query, key, and value, respectively. We compute the attention-based gating scores using the Simple Global Attention (SGA) mechanism \cite{wu2023sgformer}:
\begin{equation}
    Q_{\text{norm}} = \frac{Q}{\| Q \|_F}, \quad K_{\text{norm}} = \frac{K}{\| K \|_F}
\end{equation}

\begin{equation}
    D_g = \text{diag} \left( 1 + \frac{1}{N} Q_{\text{norm}} (K_{\text{norm}}^\top \mathbf{1}) \right).
\end{equation}
\begin{equation}
    Z = \beta D_g^{-1} \left(V + \frac{1}{N} Q_{\text{norm}} (K_{\text{norm}}^\top V) \right) + (1 - \beta) X.
\end{equation}
where \( \beta \) is a learnable residual weight controlling the contribution of the original input, and $Z$ is the attention scores for each node and experts in the pool. 

To enforce sparsity in expert selection, we introduce a learnable gating threshold \( T \in [0,1] \), which only activates experts which exceeds scores above \( T \). It is initialized as \( T_{\text{init}} \sim \{ \mathbf{0}, \mathcal{U}(0,1) \} \), where \( \mathcal{U}(0,1) \) denotes a uniform distribution over (0,1). Both \( T_{\text{init}} \) and expert scores \( Z \) are transformed via a sigmoid function to ensure values in \([0,1]\). A ReLU-based gating mask is applied for threshold comparison, the final gating scores (\( G \)) in Eq.~(\ref{Ge_eq}) uses a binary selection mask ($M$) allowing only experts above the threshold to contribute.
\begin{equation}
\label{eq:selection_mask}
    M=\operatorname{sign}(\text{ReLU}(Z' - T)). 
\end{equation}
\begin{equation}
    \text{where, } T = \sigma(T_{\text{init}}), \quad Z' = \sigma(Z)
\end{equation}
Since the sign function is non-differentiable, we customize the backpropagation by directly passing the gradient of the binary selection mask to \( \text{ReLU}(Z' - T )\), effectively bypassing the sign function. The number of activated experts (\(k_u\)) per node \(u\) is defined in Eq.~(\ref{ku_eq}).

\begin{equation}
\label{ku_eq}
    k_u = \vert {\{j \,\vert\, M_{u,j} > 0 \}} \vert. 
\end{equation}
where $M_{u,j}$ is value for node $u$ and expert $j$. For nodes where no expert is selected (\( k_u = 0 \)), we assign the expert with the highest attention score to ensure each node is allocated at least one expert: \(M_{u, \text{argmax}_j \, Z'_{u,j}} = 1\).
\begin{equation}
\label{Ge_eq}
    G = Z' \odot M.
\end{equation}

Once experts are selected, the valid experts process the input \( (X, A) \), producing expert outputs \( H_{e_j} \) for expert $e_j$, these outputs are aggregated using the gating scores in Eq (\ref{Ge_eq}). Our ablation study (Section \ref{ablation_study}) shows that the proposed gating mechanism outperforms existing alternatives, validating its effectiveness.

\begin{algorithm}[tb]
\caption{\model{} Framework (Training Phase)}
\label{alg:SAGMM}
\textbf{Input:} Graph \((X, A)\) with labels \(Y\), expert pool \(e = \{e_1, \dots, e_{N_{0}}\}\), pruning threshold \(\eta\), pruning interval \(t\).\\
\textbf{Output:} Updated expert pool \(e'\), gating weights \(G\).
 
\begin{algorithmic}[1]
\STATE \textbf{Initialize:} \(\forall\, e_i \in e,\ I(e_i) = 0\).
\STATE Divide \((X, A)\) and \(Y\) into mini-batches \(\{(X_b, A_b),\, Y_b\}\).
 
\FOR{each batch \(\bigl(X_b, A_b, Y_b\bigr)\) at epoch \(q\)}
    \STATE $X_b' \leftarrow (X_b, A_b)$  Using Eq.~(\ref{eq: toplogical_aware}).
    \STATE Obtain \(M\), \(k_u\), \(G\), and  via Eqs.~(\ref{eq:selection_mask}), (\ref{ku_eq}), (\ref{Ge_eq}), .
    \STATE Determine active experts: \(S \gets \bigcup_{u} \{\, j \,\mid\, M_{u,j} > 0 \}\).
 
    \FOR{each expert \(e_j \in S\)}
        \STATE Compute expert output \(H_{e_j} \gets e_j\bigl(X_b, A_b\bigr)\).
    \ENDFOR
 
    \STATE Aggregate expert outputs: 
    \\
    \hspace{1em} \(\bar{Y}_b \gets \sum_{j=1}^{N} G_{e_j} \,\mathrm{MLP}(H_{e_j})\).
    \STATE Evaluate loss \(\mathcal{L} \gets \mathcal{L}\bigl(Y_b,\bar{Y}_b\bigr)\) and update weights for \(e_j \in S\).
    \STATE Update expert importance \(I_t(e_i)\) via Eq.~(\ref{eq:importance_update}).
 
    \IF{\((q \bmod t) == 0\)}
        \STATE Prune experts using threshold \(\eta\): \\
        \hspace{1em} \(e \gets \{e_i \mid I(e_i) \ge \eta\}\).
    \ENDIF
\ENDFOR
 
\RETURN Updated expert pool \(e'\) and gating weights \(G\).
\end{algorithmic}
\end{algorithm}
\subsection{Adaptive Expert Pruning}
\label{section_pruning}

For efficient training and optimizing memory utilization, we also introduce an adaptive expert pruning mechanism that dynamically evaluates and updates the importance of each expert in the pool over time. 
The importance score $I(e_i)$ at pruning interval $t$ for an expert $e_i$ is updated based on its contribution in previous intervals as follows:
\begin{equation}
\label{eq:importance_update}
I_t(e_i) = (1 - \alpha) I_{t-1}(e_i) + \alpha {\gamma}(e_i).    
\end{equation}
Here, ${\gamma}(e_i)$ represents the contribution score of expert $e_i \in e$, computed as:
\begin{equation}
{\gamma(e_i)} = \left\lVert \sum_{u=1}^{n} \left( G_{e_i,u} H_{e_i,u,:} \right) \right\rVert.    
\end{equation}

where $G_{e_i,u}$ are gating weights and $H_{e_i,u,:}$ represents representation of expert $e_i$ and node $u$. Above equation captures the cumulative weighted contribution of the expert to the overall model output for the selected batch of data.

The smoothing factor $\alpha \in [0,1]$ controls the balance between previous importance scores and the current contribution, thereby preventing the premature removal of experts that may become useful later in training.
The importance scores for all experts are calculated at every epoch, and experts with importance scores below the threshold are removed at prune interval:
   \begin{equation}
   e \leftarrow \{e_i \mid I(e_i) \geq \eta\}.    
   \end{equation}

The threshold factor $\eta$, which controls the pruning aggressiveness, is dynamically adjusted based on validation performance.

\begin{figure}[!t]
    \centering
    \includegraphics[width=\linewidth]{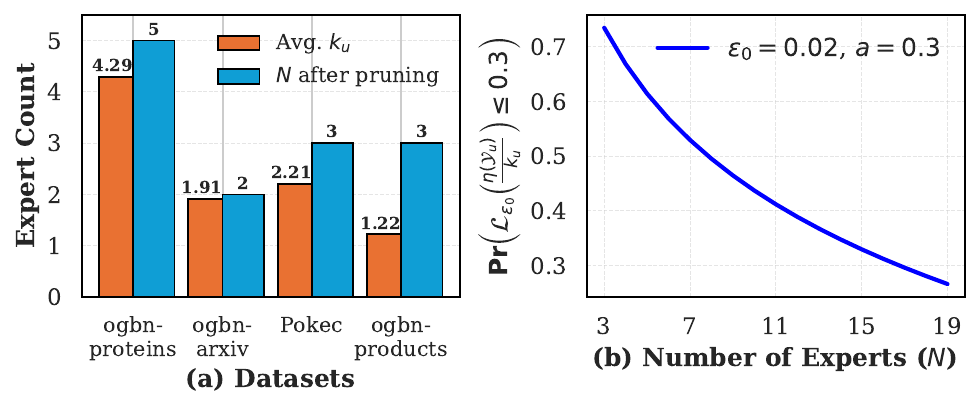}
    \caption{(a) Average number of experts activated per node $u$ ($k_u$) and experts remaining post-pruning ($N$). (b) Visualization of the Pr $\left\{ \mathcal{L}\left(\frac{\eta(\mathcal{Y}_u)}{k_u}\right) \leq a \right\}$ (Theorem 1) for $\epsilon_0 = 0.001, a = 0.3$.}
    \label{fig:side_by_side_figures_2}
\end{figure}

As shown in Figure~\ref{fig:side_by_side_figures_2}(a), the average \( k_u \) values differ from the final number of experts \textit{N} post-pruning, highlighting the dynamic and input-dependent nature of expert selection. The TAAG gating mechanism selectively activates and weighs relevant experts, while the pruning strategy removes persistently underutilized ones. This combination ensures that SAGMM maintains both specialization and computational efficiency. \\
\textbf{Auxiliary Loss Functions.}
To ensure balanced expert utilization, we incorporate importance loss and diversity loss from prior works \cite{wang2023graphmixtureexpertslearning, guo2024dynamic}. These losses prevent over-reliance on specific experts by encouraging more even and orthogonal expert activation patterns.

\subsection{Pre-trained Models as Experts}
One variant of \model{} we experimented is \model{} Pre-trained Experts (\model{}-PE), where the expert GNNs are pre-trained independently and kept fixed during the downstream task with only the router and task-specific head being trained. This design reduces computational overhead and allows efficient reuse of specialized or weak models that may not perform well individually but contribute valuable structural signals when combined. As shown in Table~\ref{table:results-table-node}, \model{}-PE achieves competitive or superior performance on node classification task. Furthermore, the results in the pre-training ratios table (see Technical Appendix) show that comparable accuracy is achieved even when the experts and the router are pre-trained on only 50\% to 70\% of the training data. This highlights the data efficiency of our framework and the robustness of its model-agnostic expert design.

\subsection {Theoretical Analysis}

\begin{table*}[!htbp]
  \centering
  
  \begin{tabular}{l|p{1.6cm}p{1.6cm}p{1.6cm}p{1.6cm}p{1.6cm}p{1.7cm}p{1.7cm}}
    \toprule
    \textbf{Model} & \textbf{Deezer} & \textbf{YelpChi} & \textbf{ogbn-proteins} & \textbf{ogbn-arxiv} & \textbf{Pokec} & \textbf{ogbn-products} & \textbf{ogbn-papers100M} \\ 
    \midrule
    GCN       & $57.70_{\pm 0.44}$ & $85.62_{\pm 0.26}$ & $69.74_{\pm 3.79}$ & $71.74_{\pm 0.29}$ & $76.52_{\pm 0.54}$ & $75.54_{\pm 0.06}$ &  $61.42_{\pm 0.25}$ \\
    GAT       & $58.59_{\pm 0.21}$ & $85.42_{\pm 0.26}$ & $69.56_{\pm 3.82}$ & $71.42_{\pm 0.09}$ & $78.87_{\pm 0.60}$ & $76.77_{\pm 0.15}$ &  $-$ \\
    GraphSAGE & $64.40_{\pm 0.79}$ & $89.23_{\pm 0.57}$ & $73.21_{\pm 1.88}$ & $71.46_{\pm 0.23}$ & $79.82_{\pm 1.62}$ & $78.29_{\pm 0.11}$ &  $63.54_{\pm 0.18}$\\
     SGC       & $57.66_{\pm 0.92}$ & $85.59_{\pm 0.25}$ & $68.75_{\pm 3.55}$ & $71.81_{\pm 0.20}$ & $76.81_{\pm 0.52}$ & $75.48_{\pm 0.14}$ &  $57.50_{\pm 0.28}$\\
    JKNet     & $64.44_{\pm 0.45}$ & $89.75_{\pm 0.32}$ & $75.67_{\pm 2.06}$ & $71.72_{\pm 0.19}$ & $78.84_{\pm 0.19}$ & $-$ &  $62.98_{\pm 0.26}$ \\
    Graph CNN & $63.65_{\pm 0.60}$ & $89.25_{\pm 0.26}$ & $77.54_{\pm 1.09}$ & $72.04_{\pm 0.19}$ & $80.21_{\pm 0.14}$ & $-$ &  $-$ \\
    GIN       & $59.71_{\pm 1.25}$ & $85.52_{\pm 0.26}$ & $54.46_{\pm 2.63}$ & $68.16_{\pm 0.27}$ & $72.53_{\pm 1.45}$ & $-$ &  $-$ \\
    MixHop    & $57.83_{\pm 1.83}$ & $85.74_{\pm 0.38}$ & $73.64_{\pm 2.34}$ & $71.94_{\pm 0.40}$ & $81.24_{\pm 0.27}$ & $-$ &  $-$\\
    \midrule
    GMoE-GCN      & $61.11_{\pm 1.19}$ & $85.75_{\pm 0.31}$ & $74.48_{\pm 0.58}$ & $71.88_{\pm 0.32}$ & $76.04_{\pm 0.14}$ & $64.18_{\pm 0.25}$ &  $60.17_{\pm 0.20}$\\
    DA-MoE    & $62.15_{\pm 0.46}$ & $85.53_{\pm 0.29}$ & $75.22_{\pm 0.16}$ & $71.96_{\pm 0.16}$ & $64.87_{\pm 5.68}$ & $68.77_{\pm 0.19}$ &  $53.63_{\pm 1.68}$\\
    \midrule
    \textbf{SAGMM} & $64.73_{\pm 0.72}$ & $91.06_{\pm 0.60}$ & $\mathbf{78.15_{\pm 1.38}}$ & $\mathbf{72.80_{\pm 0.43}}$ & $\mathbf{81.76_{\pm 0.80}}$ & $\mathbf{82.91_{\pm 0.53}}$ & $\mathbf{64.40_{\pm 0.10}}$\\
   \textbf{SAGMM-PE} & $\mathbf{65.51_{\pm 0.57}}$ & $\mathbf{91.33_{\pm 0.26}}$ & $76.28_{\pm 1.28}$ & $72.08_{\pm 0.30}$ & $80.57_{\pm 0.95}$ & $82.17_{\pm 0.47}$ &  $62.54_{\pm 0.15} $\\
    \bottomrule
  \end{tabular}
  \caption{Node classification results. For the ogbn-proteins dataset, the metric used is ROC-AUC, while testing accuracy is used for the other datasets. \textbf{SAGMM-PE}: Experts are pretrained and kept frozen, while only the router and task heads are trained. A dash (–) indicates that the corresponding model is not included in the expert pool.}
  \label{table:results-table-node}
\end{table*}
\begin{table*}[!ht]
  \centering
  
  \begin{tabular}{l|llll|ll}
    \toprule
    \multirow{2}{*}{\textbf{Model}} & \multicolumn{4}{c|}{\textbf{Graph Classification}} & \multicolumn{2}{c}{\textbf{Graph Regression}} \\
    \cmidrule{2-7}
          & \textbf{molbbbp} & \textbf{moltox21} & \textbf{moltoxcast} & \textbf{molhiv} & \textbf{molfreesolv} & \textbf{molesol} \\
    \midrule
    GCN       & $68.87_{\pm 1.51}$ & $75.29_{\pm 0.69}$ & $63.54_{\pm 0.42}$ & $76.06_{\pm 1.75}$ & $2.64_{\pm 0.24}$ & $1.11_{\pm 0.04}$ \\
    GIN       & $65.50_{\pm 1.80}$ & $74.30_{\pm 0.50}$ & $63.30_{\pm 1.50}$ & $75.40_{\pm 1.50}$ & $2.76_{\pm 0.35}$ & $1.17_{\pm 0.06}$ \\
    GraphSAGE       & $63.07_{\pm 1.83}$ & $75.64_{\pm 0.73}$ & $65.69_{\pm 0.29}$ &  $75.54_{\pm 1.28}$ & $2.43_{\pm 0.30}$ & $1.05_{\pm 0.08}$ \\
    GAT       & $67.82_{\pm 1.75}$ & $74.36_{\pm 0.70}$ & $65.50_{\pm 0.89}$ & $73.40_{\pm 1.87}$ & $2.23_{\pm 0.11}$ & $1.02_{\pm 0.05}$ \\
    \midrule
    GMoE-GCN      & $\mathbf{70.04}_{\pm \mathbf{1.12}}$ & $75.45_{\pm 0.58}$ & $64.12_{\pm 0.61}$ & $77.35_{\pm 0.63}$ & $2.50_{\pm 0.19}$ & $1.09_{\pm 0.04}$ \\
    DA-MoE    & $69.62_{\pm 0.98}$ & $75.59_{\pm 0.69}$ & $65.18_{\pm 0.48}$ & $\mathbf{77.62}_{\pm \mathbf{1.58}}$ & $2.19_{\pm 0.07}$ & $1.13_{\pm 0.04}$ \\
    \textbf{SAGMM} & $69.98_{\pm 0.18}$ & $\mathbf{76.58}_{\pm \mathbf{0.88}}$ & $\mathbf{66.63}_{\pm \mathbf{0.53}}$ & $77.48_{\pm 1.10}$ & $\mathbf{2.15}_{\pm \mathbf{0.12}}$ & $\mathbf{0.93}_{\pm \mathbf{0.03}}$ \\
    \bottomrule
  \end{tabular}
  \caption{Graph classification and regression results. Metric used is testing ROC-AUC for classification and RMSE for regression. Best model per column is shown in bold.}
  \label{results-table-graph}
\end{table*}
 We provide a theoretical justification for using expert mixtures and pruning in SAGMM by analyzing how sparsely activating a subset of experts affects model performance.\\
\noindent
Consider a simplified setting where 

(1) The GNN has a single-layer architecture with an update rule defined in Eq.~\ref{forwardprop} for binary node classification. (2) Input features $X = [x_1, x_2, \ldots, x_n]$, where $x_i \sim \mathcal{N}(0_d, \sigma^2 I_{d \times d})$ with $k_u > 2$ experts for each node $u$. (3) The aggregated output at layer 1 for node \( u \) is given by  
\(
\frac{\eta(\mathcal{Y}_u)}{k_u},
\)
where \( \mathcal{Y}_u = \sum_{m=1}^{k_u} \hat{H}^{(1)}_{e_i,u} \). Here, \( \hat{H}^{(1)}_{e_i, u} \) represents the first-order approximation of \( H^{(1)}_{e_i, u} \) (Details in Technical Appendix). The function \( \eta(x) \) is defined as  
\(
\eta(x) = \frac{\exp(\boldsymbol{1}^T x)}{1 + \exp(\boldsymbol{1}^T x)}.
\) (4) The loss function $\mathcal{L}$ is the numerically stable binary cross-entropy loss:
    $$\mathcal{L}_{\epsilon_0}(x, y_u) = -y_u \log(x + \epsilon_0) - (1 - y_u)\log(1 - x + \epsilon_0).$$

\begin{theorem}
Under the design conditions above, the following inequality holds:
\begin{equation}
\Pr\left\{ \mathcal{L}_{\epsilon_0}\left( \frac{\eta(Y_u)}{k_u} \right) \leq a \right\}
\leq \frac{U - f(k_u, \epsilon_0)}{U - a},
\end{equation}
where $U = -\log(\epsilon_0)$ and 
$
f(k_u, \epsilon_0) = \log(2k_u^2) - \log(2k_u(1 + \epsilon_0) - 1).
$
\end{theorem}

This bound reveals that increasing the number of activated experts $k_u$ generally improves the tightness of the bound, lowering expected error. However, as shown in Figure~\ref{fig:side_by_side_figures_2}(b), pruning a small number of low-importance experts only marginally impacts the bound while substantially reducing computation. This result supports our adaptive pruning mechanism in SAGMM. A detailed proof and a more general version of this theorem are provided in Technical Appendix.

\section{Experiments}
\label{experiments}

\noindent \textbf{Initial Number of Experts ($N_{0}$).} For our primary task of node classification, we construct the expert pool by selecting four and eight widely recognized GNN architectures: GCN, JKNet, GraphCNN, MixHop, GAT, SGC, GIN, and GraphSAGE, all of which are established standards for this application \cite{hu2020open}. In the case of other tasks, the expert pool is composed of four GNN models. The decision to use either eight or four experts is guided by the need to balance architectural diversity with computational feasibility. Notably, this choice is consistent with recent findings in the literature \cite{yun2024toward, he2025efficiently}, which indicate that configuring MoE models with four or eight experts enables efficient inference and strong model performance. 
\\
\begin{table}[!ht]
  \centering
  
  \begin{tabular}{l|lll}
    \toprule
    \textbf{Model} & \textbf{ogbl-ddi} & \textbf{ogbl-collab} & \textbf{ogbl-ppa}  \\
    \midrule
    GraphSAGE      & $53.90_{\pm 4.74}$  & $48.10_{\pm 0.81}$  & $16.55_{\pm 2.40}$  \\
    GCN            & $37.07_{\pm 5.07}$           & $44.75_{\pm 1.07}$  & $18.67_{\pm 1.32}$ \\
    JKNet          & $60.56_{\pm 8.69}$         & $51.47_{\pm 0.59}$  & \multicolumn{1}{c}{$21.58_{\pm 2.59}$} \\
    SGC          & $35.00_{\pm 4.65}$         & $48.99_{\pm 0.60}$  & \multicolumn{1}{c}{$9.64_{\pm 1.60}$} \\
    \midrule
    GMoE-GCN          & $37.96_{\pm 8.20}$          & $32.61_{\pm 2.59}$  & $19.25_{\pm 1.67}$   \\
    DA-MoE         & $45.58_{\pm 6.93}$           & $47.64_{\pm 3.77}$ & ${21.46}_{\pm 2.53}$ \\
    \textbf{SAGMM} & $\mathbf{74.20}_{\pm 11.71}$   & $\mathbf{52.39_{\pm 0.40}}$ &  $\mathbf{26.11_{\pm 1.12}}$ \\
    \bottomrule
  \end{tabular}
  \caption{Link prediction results. Metric used is HITS@20 for ogbl-ddi, HITS@100 for ogbl-ppa and HITS@50 for ogbl-collab as per OGB benchmark protocol.}
  \label{results-table-link}
\end{table} 
\textbf{Baselines.} For each of the tasks, we compare against two types of baselines: 1) existing SOTA graph-based MoE architectures: GMoE and DA-MoE, and 2) each of the individual expert in the pool.\\
\textbf{Settings.} We use a common set of hyperparameters across all methods to ensure consistency. For model-specific settings, we adopt the default configurations provided in the respective original papers. All reported results are averaged over ten independent runs to ensure a fair comparison. A detailed description of the hyperparameter settings is provided in the Technical Appendix.

\label{results}

\subsection{Performance Evaluation}
We evaluate our framework across three categories of tasks: node classification, graph-level prediction, and link prediction, using a broad set of benchmark datasets.
\\
\textbf{Node Classification.} We evaluate on seven standard datasets ranging from small to large-scale, including four homophilic graphs: \textit{ogbn-arxiv}, \textit{ogbn-proteins}, \textit{ogbn-products}, and \textit{ogbn-papers100M} from the OGB benchmark \cite{hu2020open}, and three heterophilic graphs: \textit{Pokec} \cite{jure2014snap}, \textit{Deezer} \cite{rozemberczki2020characteristic}, and \textit{YelpChi} \cite{mukherjee2013yelp}. As shown in Table~\ref{table:results-table-node}, \model{} outperforms all baselines, achieving notable gains on medium to large-scale datasets such as a 5.90\% improvement on \textit{ogbn-products} and 1.57\% on \textit{ogbn-papers100M}, along with consistent accuracy improvements across the remaining benchmarks. Our analysis reveals that \model{} consistently prunes weaker experts such as GIN across datasets, demonstrating its ability to adaptively select the most effective experts.
\\
\textbf{Graph-Level Prediction.} We use six molecular property prediction datasets from the OGB benchmark: \textit{molhiv}, \textit{moltox21}, \textit{moltoxcast}, \textit{molbbbp}, \textit{molesol}, and \textit{molfreesolv}. As shown in Table \ref{results-table-graph}, \model{} achieves strong performance, with notable gains on \textit{moltox21}, \textit{moltoxcast}, and \textit{molesol}. It incurs minor drops on \textit{molhiv} and \textit{molbbbp}, but remains competitive overall.
\\
\textbf{Link Prediction.} We evaluate \model{} on three standard OGB link prediction datasets: \textit{ogbl-ppa}, \textit{ogbl-ddi}, and \textit{ogbl-collab}. As shown in Table~\ref{results-table-link}, \model{} achieves substantial performance improvements, with gains of 22.52\% on \textit{ogbl-ddi}, 20.90\% on \textit{ogbl-ppa}, and 1.78\% on \textit{ogbl-collab}, outperforming all baselines across the board. \\

\textbf{Inference Time and Memory Usage.} As shown in Table \ref{time-gpu-memory-table}, SAGMM maintains competitive inference time and GPU memory usage compared to existing methods. Additionally, aggressive expert pruning can further reduce inference cost, with only a minor trade-off in performance, enabling a controllable balance between efficiency and effectiveness.

\begin{table}[h]
  \centering
  
  \begin{tabular}{l|cc|cc}
    \toprule
    \multirow{3}{*}{\textbf{Model}} & \multicolumn{2}{c|}{\textbf{ogbn-arxiv}} & \multicolumn{2}{c}{\textbf{ogbl-ddi}} \\
    \cmidrule{2-5}
                                    & Time           & GPU Mem       & Time           & GPU Mem       \\
                                    & (ms)           & (GB)             & (ms)           & (GB)             \\
    \midrule
    GCN                             & 126             & 1.74             & 235            & 1.50             \\
    GMoE                            & 200             & 3.73             & 298            & 1.50             \\
    DA-MoE                          & 251             & 1.90             & 361            & 2.15             \\
    \midrule
    \textbf{SAGMM}                           & 227            & 5.89             & 334            & 1.95             \\
    \bottomrule
  \end{tabular}
  \caption{Inference time and GPU memory usage comparison across models on ogbn-arxiv and ogbl-ddi datasets.}
  \label{time-gpu-memory-table}
\end{table}

\subsection{Ablation Study}
\label{ablation_study}

\begin{table}[!ht]
\centering

\setlength{\tabcolsep}{3pt} 
\begin{tabular}{l|cc}
\toprule
\textbf{Method} & \textbf{ogbn-proteins} & \textbf{ogbl-collab} \\
\midrule

\textbf{SAGMM} & $\mathbf{78.15}_{\pm \mathbf{1.38}}$  & $ 52.39_{\pm 0.40}$ \\
SAGMM \emph{(w/o diverse)} & $69.89_{\pm 1.84}$ &  $40.08_{\pm 4.73}$ \\
SAGMM \emph{(top-k gating)} & $67.29_{\pm 3.94}$ & $48.67_{\pm 0.94}$\\
SAGMM \emph{(top-any gating)} & $77.31_{\pm 2.31}$ & $50.42_{\pm 0.99}$\\
SAGMM \emph{(w/o pruning)} & $75.24_{\pm 1.49}$ & $\mathbf{52.44}_{\pm \mathbf{0.57}}$ \\
\bottomrule
\end{tabular}
\caption{Results of ablation study. Metrics are ROC-AUC for ogbn-proteins and HITS@50 for ogbl-collab.}
\label{ablation-removing-components}
\end{table}
We conduct an ablation study to evaluate the contribution of each component in SAGMM by testing four variants: (i) using identical GNN architectures for all experts (\textit{SAGMM (w/o diverse)}), (ii) replacing our gating module with noisy Top-$k$ gating (\textit{SAGMM (top-$k$ gating)}), (iii) using Top-Any gating~\cite{guo2024dynamic} having dynamic expert selection (\textit{SAGMM (top-any gating)}), and (iv) disabling the adaptive expert pruning mechanism (\textit{SAGMM (w/o pruning)}). As shown in Table~\ref{ablation-removing-components}, the largest drop is observed when expert diversity is removed, confirming its critical role. Substituting our gating mechanism with top-$k$ or top-any variants impairs expert selection quality, while disabling pruning retains performance in the case of \textit{ogbl-collab} but always leads to higher memory usage. These results underscore the importance of expert diversity, TAAG gating, and pruning in SAGMM's effectiveness.

\section{Conclusion}
\label{conclusion}
This paper presents SAGMM, a novel framework that employs multiple GNN models as experts within a MoE paradigm. It features a topology-aware attention gating mechanism tailored for graph data and an adaptive pruning strategy. Extensive experiments demonstrate that \model\ consistently outperforms traditional approaches across diverse downstream tasks. While \model\ currently focuses on GNN-based experts, in principle, it can integrate any black-box model. 
\\
Future work will explore extending SAGMM through expert distillation, where the knowledge from a diverse pool of experts and the TAAG gating mechanism is transferred into a compact, unified student model. Also, SAGMM for dynamic graphs is another promising direction for further research.

\bibliography{main}

\include{appendix}

\end{document}

%% file: appendix.tex
\onecolumn


\appendix
\begin{center}
\Huge \textbf{Technical Appendix}
\end{center}

\section{Case Study}
The effectiveness of the SAGMM method is demonstrated through a meticulously designed scenario below. As shown in \cite{xu2018powerful}, GNNs can be at most as powerful as the WL test where the input to the GNN is considered as $\boldsymbol{1}$ or $C \boldsymbol{1}$, where $C$ is the graph operator. For the SAGE \cite{hamilton2017inductive} model the graph operator, $C_{\text{\tiny SAGE}} = I_n + D^{-1}A$, with $D$ representing the degree matrices of the graph.

In this setup, we consider three experts: GCN \cite{kipf2016semi}, SAGE, and GIN \cite{xu2018powerful} — under all one's input as that of the WL test. Among these experts, GIN emerges as the preferred choice, where the discriminatory power of GIN is shown to be as powerful as the WL-test \cite{xu2018representation}. We demonstrate a gating function for the SAGMM model that effectively selects the GIN model under all one's input for the first layer. The outputs of these experts at the first layer are given by:
$
H_{\text{GIN}}^{(1)} = \text{MLP}^{(1)}\left((A + (1+\epsilon^{(l)})I_n )\boldsymbol{1}\right), \;
H_{\text{GCN}}^{(1)} = \sigma \left( F^{-1} \boldsymbol{1} \Theta^{(0)} \right), \;
H_{\text{SAGE}}^{(1)} = \sigma \left( 2 \boldsymbol{1} \Theta^{(0)} \right).
$
$F$ is a diagonal matrix with entries $F_{ii} = \sum_j C_{\text{\tiny GCN}}(ij)$ 
The gating function for SAGMM is detailed in Algorithm 1 in main paper, but for brevity, let us consider a gating score function defined for an expert $e_i \in \{\text{GCN, SAGE, GIN}\}$ and batch $b$ as
\begin{equation}
\label{eq:case_study}
 \alpha_{b,e_i} = \sum_j \text{Var}\left(Y_{e_i}(:,j)\right), \quad Y_{e_i}(p,:) = \frac{H^{(1)}_{e_i}(p,:)}{H^{(1)}_{e_i}(p,:) \boldsymbol{1}_{d\times 1}}   
\end{equation}
 In Eq.~(\ref{eq:case_study}),  Var denotes the variance operation. It can be observed that the gating scores for GCN, SAGE will be zero. The ideal architecture in this case, GIN will be selected for which the variance will be non-zero. The advantage of the SAGMM model described in the current scenario is distinct from prior works on a mixture of experts, such as \cite{wang2023graphmixtureexpertslearning,yao2024moe}, where the model relies on a single model type and applies MoE within intermediate layers. The selected model type in these prior works must be GIN to achieve optimal performance.
\section{Theoretical Insights}
\subsection{Background}
\subsubsection{Count Sketch.}  
\label{countsketch}
Matrix multiplication plays a fundamental role in machine learning and scientific computing, with numerous optimization techniques explored in prior works such as \cite{pytorchamm}, \cite{guennebaud2010eigen}, and \cite{abadi2016tensorflow}. Count sketch, a powerful dimensionality reduction method introduced in \cite{Count-sketch} maps an \( n \)-dimensional vector \( u \) to a lower-dimensional space of size \( c \). This transformation utilizes a random hash function \( h: [n] \rightarrow [c] \) and a Rademacher variable \( s: [n] \rightarrow \{-1,1\} \). The dimensionality reduction process is defined as  
\[
CS(u)_i = \sum_{h(j)=i} s(j) u_j = R(i, :) u,
\]  
where \( R \in \mathbb{R}^{c \times n} \) represents the count sketch matrix.

\subsubsection{Tensor Sketch.}  
\label{tensorsketch}  
Tensor sketch extends the concept of count sketch \cite{Count-sketch} and is widely used for efficient dimensionality reduction in large-scale machine learning applications \cite{Tensorsketchfirst}. Pham and Pag \cite{Phamandpag} introduced an efficient implementation leveraging the Fast Fourier Transform (FFT) and its inverse, formulated as  
\[
\text{TS}_k(A) = \text{FFT}^{-1} \bigodot_{p=1}^k \text{FFT} \left( \text{CS}^{(p)} (A) \right).
\]  
In \cite{ding2022sketchgnn}, tensor sketching is applied to approximate the element-wise \( k \)-th power of a matrix product:  
\[
(AB)^{\odot k} \approx \text{TS}_k(A) \, \text{TS}_k(B^T)^T.
\]  
Here, \( A \in \mathbb{R}^{n \times n} \) and \( B \in \mathbb{R}^{n \times d} \), while their tensor sketch representations satisfy \( \text{TS}_k(A) \in \mathbb{R}^{n \times c} \) and \( \text{TS}_k(B^T) \in \mathbb{R}^{d \times c} \), with \( c \) chosen such that \( c < n \).
\\
 To study the impact of pruning experts during training, we investigate the loss of the GNN under diverse mixtures of experts. Towards this, we present Theorem 1 (as listed in main paper also), where we investigate the GNN with a mixture of experts considering randomized inputs to the GNN. The input feature matrix to the GNN considered are \( X = [x_1, x_2, \ldots, x_n] \), where each feature \( x_u \) is 
independently sampled from a multivariate normal distribution, 
\( \mathcal{N}(\mathbf{0}_d, I_{d \times d}) \). Let the experts be given by $\{e_1, e_2\ldots e_N\}$. The investigation of GNN properties using such randomized inputs is inspired by \cite{kanatsoulis2022representation}. The article demonstrates how randomized inputs for the input feature matrix are a more effective approach for analyzing GNN behaviour compared to the inputs used in the WL test.
\subsection{Assumptions of Theorem 1}
\label{assumption_gnn}
Theorem 1 is based on the following assumptions:
\begin{itemize}
    \item The GNN has a single-layer architecture with an update rule defined in Eq.~(\ref{forwardprop}) with the downstream task of binary node classification.
    \item Input features $X = [x_1, x_2, \ldots, x_n]$, where $x_i \sim \mathcal{N}(0_d, \sigma^2 I_{d \times d})$ with the number of experts for every node $u$, $k_u > 2$.
    \item The aggregated output at layer 1 for node $u$ is $\frac{\eta(\mathcal{Y}_u)}{k_u}$. $\mathcal{Y}_u = \sum_{m=1}^{k_u} H^{(1)}_{e_i u}$. The posterior function $\eta(x) = \frac{\exp(\boldsymbol{1}^T x)}{1 + \exp(\boldsymbol{1}^T x)}$.
    \item The loss function ($\mathcal{L}$) is the numerically stable binary cross-entropy loss $$\mathcal{L}_{\epsilon_0}(x, y_i) = -y_i \log(x + \epsilon_0) - (1 - y_i)\log(1 - x + \epsilon_0).$$  
\end{itemize}
\begin{figure*}[htbp]
  \centering
  \includegraphics[width=0.7\textwidth]{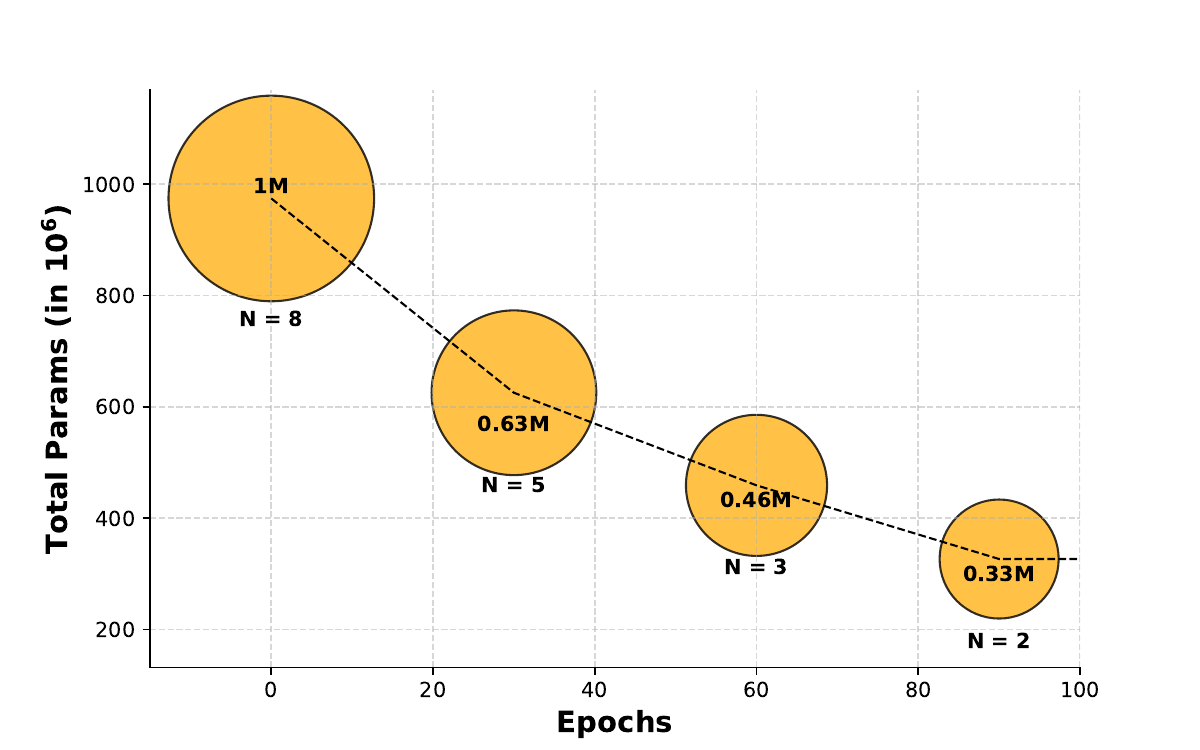} 

  \caption{Paramaters reduction with expert pruning on ogbn-arxiv dataset. M denotes Millions.}
  \label{fig:arxiv_paramter_reduction}
\end{figure*}
\begin{theorem}
\label{main_theorem}
Under the design conditions specified in Assumption above, the following inequality holds:
\begin{equation}
    \text{Pr}\left\{\mathcal{L}_{\epsilon_0}\left(\frac{\eta(\mathcal{Y}_u)}{k_u}\right) \leq a \right\} \leq \frac{U - f(k_u, \epsilon_0)}{U - a}.
\end{equation}
$f(k, \epsilon_0) = \log(2k_u^2) - \log(2k_u(1+\epsilon_0) - 1)$, $U = -log(\epsilon_0)$, $a > 0$ is a positive constant. 
\end{theorem}
\noindent This result highlights how a first-order approximation of the non-linear activation in GNNs for a binary node classification task benefits from a mixture of experts. In the SAGMM method, we employ an \emph{expert pruning} mechanism, which progressively reduces the number of experts during training, as illustrated in Figure \ref{fig:arxiv_paramter_reduction}. From Figure 3(b) in main paper, we observe that the upper bound for the probability of loss (Pr $\left\{ \mathcal{L}\left(\frac{\eta(\mathcal{Y}_u)}{k_u}\right) \leq a \right\}$) for $\epsilon_0 = 0.001, a = 0.3$ decreases with more experts. However, reducing the number of experts from 8 to 6 results in only a minor decrease in the upper bound, approximately 0.075, while also reducing the number of parameter updates required which improves the training efficiency.
\paragraph{ \textbf{{Proof of Theorem 1}}}

The proof builds upon the approximation of GNN update rules in Eq.~(\ref{forwardprop}) as introduced in \cite{ding2022sketchgnn}. The non-linearity in the update rule is approximated using a power series expansion. Tensor-Sketch \cite{tensorsketch} is then applied to approximate the series. For a single convolution matrix $C^{(l)}$ with expansion order $r$ we have the following,
$$
H^{(l+1)} = \sigma(C^{(l)} H^{(l)} W^{(l)}) \approx \sum_{k=1}^r c_k (C^{(l)} H^{(l)} W^{(l)})^{\odot r}  \approx \sum_{k=1}^r c_k \text{TS}_k(C^{(l)}) \text{TS}_k\left(\left(H^{(l)} W^{(l)} \right)^T\right)^T.
$$
$c_k$ refers to the learnable coefficients that integrate representations of different powers. 
\\
\textbf{Notation:} We denote the node representations at layer $1$ for node $u$ and expert $e_i$ using a first-order approximation as follows:
\[ \hat{H}^{(1)}_{e_i u} = c_{1 e_i} \text{TS}_1(C_{e_i u,:}^{(0)}) \,\text{TS}_1\left(\left(H_{e_i}^{(0)} W_{e_i}^{(0)} \right)^T\right)^T = A_{1e_i u} \,H^{(1)}_{e_i} W^{(0)}_{e_i}, \]
where the matrix \( A_{1 e_i u} \in \mathbb{R}^{1\times n} = c_{1e_i} \text{TS}_1(C_{e_i u,:}^{(0)})({R^{(1)}})^T \). Here, $R^{(1)} \in \mathbb{R}^{n \times c}$ is the count-sketch matrix ~\cite{tensorsketch}, with $c$ representing a dimension smaller than $n$.

\begin{proof}
Applying the Reverse Markov inequality \cite{abhishek2019introduction} to the random variable $\mathcal{L}(\frac{\eta(\mathcal{Y}_u)}{k_u})$, we obtain:
$$
\text{Pr}\left\{\mathcal{L}\left(\frac{\eta(\mathcal{Y}_u)}{k_u}\right) \leq a \right\} \leq \frac{U - \mathbb{E}(\mathcal{L}(\frac{{\eta(\mathcal{Y}_u)}}{k_u}))}{U - a}.
$$
\[\mathcal{L}\left(\frac{\eta(\mathcal{Y}_u)}{k_u}, y_{u}, \epsilon_0\right) = -y_{u} \log\left(\eta(\mathcal{Y}_u) + \epsilon_0\right) - (1-y_{u}) \log\left(k_u - \eta(\mathcal{Y}_u) + \epsilon_0\right) + \log(k_u).\]
\\
By reformulating, we obtain:
\[\begin{aligned}
\mathcal{L}\left(\frac{\eta(\mathcal{Y}_u)}{k_u}, y_{u}, \epsilon_0\right) &= -y_{u} \log\left(e^{\boldsymbol{1}^T \mathcal{Y}_u}(1 + \epsilon_0) + 1\right) - (1 - y_{u}) \log\left(e^{\boldsymbol{1}^T \mathcal{Y}_u}(k_u(1 + \epsilon_0) - 1) + k_u(1 + \epsilon_0)\right) \\
&\quad + y_{u} \log\left(k_u (1 + e^{\boldsymbol{1}^T \mathcal{Y}_u})\right) + (1 - y_{u}) \log\left(k_u (1 + e^{\boldsymbol{1}^T \mathcal{Y}_u})\right) + \log(k_u).
\end{aligned}\]
\\
By applying Jensen's inequality \cite{boyd2004convex}, since $\mathcal{L}$ is a convex function of $\mathcal{Y}_u$ (as log-sum-exp, -log is convex), and using the fact that $\mathbb{E}(\mathcal{Y}_u) = 0$, we obtain:
\begin{align*}
\mathbb{E}\left(\mathcal{L}\left(\frac{\eta(\mathcal{Y}_u)}{k_u}, y_{u}, \epsilon_0\right)\right) 
& \geq \log(2k_u^2) - \log\left(2k_u(1 + \epsilon_0) - 1\right)  = f(k_u, \epsilon_0).
\end{align*}
\\
Using the lower bound of the expectation obtained above in the  Reverse Markov's inequality, we obtain:
\begin{align*}
\text{Pr}\left\{\mathcal{L}\left(\frac{\eta(\mathcal{Y}_u)}{k_u}\right) \leq a \right\} 
& \leq \frac{U - f(k_u, \epsilon_0)}{U - a}.
\end{align*}

\end{proof}

\section{Dataset and Experiment Settings}
    \subsection{Datasets Description}
    
    \paragraph{Node classification:}
    \begin{itemize}
        \item \textbf{Deezer-Europe} \cite{rozemberczki2020characteristic}: It is a user-user network of European users of the Deezer music streaming platform. Nodes represent individual users, and edges denote mutual friendships between them. Node features are derived from the artists liked by these users. The task is to classify users' gender. Following the benchmark setup in \cite{lim2021newbenchmarkslearningnonhomophilous}, the dataset is split randomly into 50\%/25\%/25\% for training, validation, and testing.
        \item \textbf{YelpChi} \cite{mukherjee2013yelp}: This dataset is a collection of user reviews from Yelp for businesses (hotels and restaurants) located in the Chicago area. The dataset contains both filtered (suspected fake) and unfiltered (genuine) reviews as identified by Yelp’s internal spam filter. The task is to detect deceptive reviews using a binary classification setting, where filtered reviews are treated as fake and unfiltered ones as non-fake . The dataset follows the real-world class imbalance naturally present in Yelp's data.
        \item \textbf{ogbn-proteins}: This is a graph where nodes represent proteins, and edges indicate biologically significant associations, categorized by type based on species.  The dataset contains proteins from 8 species. The task is a multi-label binary classification to predict the presence or absence of 112 protein functions. The public OGB \cite{hu2020open} split  is used.
        \item \textbf{ogbn-arxiv}: This dataset is a citation network of Computer Science (CS) papers on arxiv. Nodes represent papers, and edges indicate citation relationships. Each node is associated with a 128-dimensional feature vector derived by averaging the word embeddings of the paper's title and abstract, generated using the WORD2VEC model. The objective is to predict the subject area of each paper, covering 40 distinct subject areas. The dataset follows the split defined in \cite{hu2020open}.
        \item \textbf{Pokec} \cite{jure2014snap}: It is a large-scale social network dataset containing various profile features, such as geographical region, registration time, and age. The task is to predict users' gender based on these features. The dataset is split randomly into 50\% for training, 25\% for validation, and 25\% for testing.
        \item \textbf{ogbn-products}: Itis based on the Amazon Co-Purchasing network \cite{mcauley2015inferringnetworkssubstitutablecomplementary}, where nodes represent products, and edges signify frequent co-purchases between products. Node features are created using bag-of-words representations of product descriptions, and the labels correspond to the top-level categories of the products. The dataset follows public OGB split.
        \item \textbf{ogbn-papers100M}: This dataset is a directed citation graph of 111 million papers indexed by MAG \cite{10.1162/qss_a_00021}. The task is to predict the subject areas of the subset of papers that are published in arXiv. The dataset follows public OGB split.

    \end{itemize}

    \paragraph{Link prediction:}
    \begin{itemize}
        \item \textbf{ogbl-ppa}: This dataset is an undirected, unweighted graph. Nodes represent proteins from 58 different species, and edges indicate biologically meaningful associations between proteins, e.g., physical interactions, co-expression, homology or genomic neighborhood. Each node contains a 58-dimensional one-hot feature vector that indicates the species that the corresponding protein comes from. The task is to predict new association edges given the training edges. The evaluation is based on how well a model ranks positive test edges over negative test edges. It follows given OGB  split.
        \item \textbf{ogbl-collab}: This dataset is an undirected graph, representing a subset of the collaboration network between authors indexed by MAG. Each node represents an author and edges indicate the collaboration between authors. All nodes come with 128-dimensional features, obtained by averaging the word embeddings of papers that are published by the authors. All edges are associated with two meta-information: the year and the edge weight, representing the number of co-authored papers published in that year.The task is to predict the future author collaboration relationships given the past collaborations. The goal is to rank true collaborations higher than false collaborations.
        \item \textbf{ogbl-ddi}: This dataset is a homogeneous, unweighted, undirected graph, representing the drug-drug interaction network. Each node represents an FDA-approved or experimental drug. Edges represent interactions between drugs and can be interpreted as a phenomenon where the joint effect of taking the two drugs together is considerably different from the expected effect in which drugs act independently of each other. The task is to predict drug-drug interactions given information on already known drug-drug interactions. The dataset does not contain any node features, a random tensor is created following the OGB guidelines.
    \end{itemize}
    
    \paragraph{Graph classification \& Regression:}
    \begin{itemize}
        \item \textbf{ogbg-mol* (Molecular Graphs)}: The ogbg-molhiv and ogbg-molpcba datasets are two molecular property prediction datasets of different sizes: ogbg-molhiv (small) and ogbg-molpcba (medium). Each graph represents a molecule, where nodes are atoms, and edges are chemical bonds. Input node features are 9-dimensional, containing atomic number and chirality, as well as other additional atom features such as formal charge and whether the atom is in the ring. Input edge features are 3-dimensional, containing bond type, bond stereochemistry as well as an additional bond feature indicating whether the bond is conjugated. The task is to predict the target molecular properties as accurately as possible, where the molecular properties are cast as binary labels. The dataset follows OGB \cite{hu2020open} split.
    \end{itemize}
    Table \ref{tab:info-table-node} and \ref{tab:info-table-graph} summarizes the key statistics of all datasets used in our experiments.
    \begin{table}[htbp]
    
    \centering
    \begin{tabular}{lrrrrr}
    \toprule
    \textbf{Dataset} & \textbf{Number of Nodes} & \textbf{Number of Edges} & \textbf{Feature Dim} & \textbf{Number of Classes} \\
    \midrule
    Deezer             & 28,281     & 92,752        & 31,241  & 2   \\
    YelpChi            & 45,954     & 3,846,979     & 32      & 2   \\
    ogbn-proteins      & 132,534    & 39,561,252    & 8       & 112 \\
    ogbn-arxiv         & 169,343    & 1,166,243     & 128     & 40  \\
    Pokec              & 1,632,803  & 15,311,282    & 65      & 2   \\
    ogbn-products      & 2,449,029  & 61,859,140    & 100     & 47  \\
    ogbn-papers100M    & 111,059,956& 1,615,685,872 & 128     & 172 \\
    \midrule
    ogbl-ddi           & 4,267      & 1,334,889     & --      & --  \\
    ogbl-collab        & 235,868    & 1,285,465     & 128      & --  \\
    ogbl-ppa           & 576,289    & 30,326,273    & 58      & --  \\
    \bottomrule
    \end{tabular}
    \caption{Statistics of datasets used for node classification and link prediction.}
    \label{tab:info-table-node}
    \end{table}
    \begin{table}[htbp]
    
    \centering
    \begin{tabular}{lrrrrr}
    \toprule
    \textbf{Dataset} & \textbf{Number of Graphs} & \textbf{Nodes per Graph} & \textbf{Edges per Graph} & \textbf{Feature Dim} \\
    \midrule
    ogbg-molbbbp      & 2,039   & 24.6  & 25.6 & 9   \\
    ogbg-moltox21     & 7,831   & 18.5  & 19.3 & 9   \\
    ogbg-moltoxcast   & 8,576   & 18.8  & 19.2 & 9   \\
    ogbg-molhiv       & 41,127  & 25.5  & 27.5 & 9   \\
    \midrule
    ogbg-molfreesolv  & 642     & 8.7   & 8.4  & 9       \\
    ogbg-molesol      & 1,123   & 13.3  & 13.7 & 9       \\
    \bottomrule
    \end{tabular}
    \caption{Statistics of datasets used for graph classification and regression tasks.}
    \label{tab:info-table-graph}
    \end{table}

    \subsection{System Configuration}
    All experiments were conducted on a server equipped with two 64-core Intel Xeon Platinum 8562Y+ CPUs with 512GB memory. 
    The platform runs on Ubuntu 22.04.5 LTS with GCC version 10.5.0. 
    We used CUDA 11.8 and pytorch version 2.1.2 for all the experiments. Most experiments were performed on a single NVIDIA A40 GPU with 44,GB of VRAM. For large-scale datasets such as ogbn-papers100M and ogbn-products, experiments were conducted on an NVIDIA H100 GPU.
    

    \subsection{Experiment Settings}
    
    
    \textbf{Training Details}. This architecture leverages the strengths of multiple GNN designs while dynamically adapting to the input graph through the gating mechanism, enabling robust and versatile learning.
    The datasets used in our experiments encompass both heterophilic and homophilic graphs. For graphs that fit entirely within GPU memory, we employ full-graph training, while for larger graphs that exceed GPU memory limits, we adopt a mini-batch training scheme.\\
    For the datasets listed in Table \ref{tab:info-table-node} and Table \ref{tab:info-table-graph}, full-graph training is employed for ogbn-arxiv and yelpchi in the node classification task, ogbl-ddi and ogbl-ppa in link prediction, and for all datasets in graph classification and regression tasks.  In this approach, the entire graph is provided as input to the model during both training and inference. During training, the model processes the full graph to compute forward passes and predict labels for loss computation. Similarly, during inference, the entire graph is inputted, and metrics are computed based on predictions for validation and test nodes.\\
    For larger datasets, where the graph sizes prevent full-graph training, we utilize a mini-batch training approach similar to the method in \cite{wu2023nodeformerscalablegraphstructure}. At the beginning of each training epoch, the nodes are shuffled and partitioned into random mini-batches of size B. During each iteration, one mini-batch is selected, and the corresponding subgraph is extracted from the original graph. This subgraph is fed into the model to compute the loss on the training nodes within that batch. For ogbn-products and ogbn-papers100M we employ neighbor sampling training approach. \\
    To evaluate model performance, we use the model that achieves the highest accuracy on the validation set. Each experiment is conducted over ten independent trials with different initializations. Training is performed for 1000 epochs by default , with exception made for larger datasets: 150 epochs for ogbn-papers100M, 500 epochs for ogbn-products and 100 epochs for ogbl-ppa. We report the mean and variance of the evaluation metrics across these runs.
    
   \noindent \newline \textbf{Hyperparameters.} The common hyperparameters used for training are summarized in Table~\ref{tab:model_paramters}. In addition, the following gating mechanism configurations were consistently applied across all GNN models and datasets:

        \begin{itemize}
            \item \textbf{Prune type} $\in$ \{ raw\_gates, thresholded\_gates \}:  
            raw\_gates uses the raw gate scores given by router to determine expert pruning, while thresholded\_gates applies experts threshold to the gate scores which tells to consider only the activated experts for pruning.
            \item \textbf{Gate initialization} $\in$ \{ zeros, randn \}:  
            Gate values are initialized either with all zeros or drawn from a normal distribution. Zero initialization yields uniform soft selection ($\sigma(0) = 0.5$), assigning equal importance to all experts initially.

            \item \textbf{Pruning interval:} Selected from the range 20–40, depending on the dataset.
             
            \item \textbf{Auxilary loss weights} $\in$ \{ 0.05, 0.1 \}:  
            Both importance loss weight and diversity loss weight are tuned from this shared range across all datasets.
            \item \textbf{Fan out:} 15,10,5 for ogbn-papers100M and 10,5,2 for ogbn-products 
            \item \textbf{Global Information for Graph-level Tasks}: For graph classification and regression tasks, where graphs often contain a small number of nodes (e.g., fewer than 20), we additionally experimented with Graph Mean and its Laplacian-based variants to incorporate global structural information into the gating mechanism.
            \item \textbf{Weight decay: } $1 \times 10^{-4}$ for both ogbn-papers100M and ogbn-products datasets.
            \item \textbf{Optimizer:} Adam optimizer with dataset-specific learning rates.
            \item \textbf{Importance threshold factor} $\in$ \{ 0.3 - 0.7 \}:  
            It determines the pruning threshold
        \end{itemize}
    
        \begin{table}[htbp]
        \centering
            \begin{tabular}{lccccccc}
                \toprule
                Dataset & LR & Hidden & Layers & Dropout & Batch Size  \\
                \midrule
                Deezer           & 0.01  & 96 & 3 & 0.4 & 10000   \\
                YelpChi          & 0.01  & 64 & 3 & 0.5  & -   \\
                ogbn-proteins    & 0.01 & 64 & 3 & 0.0 & 50000   \\
                ogbn-arxiv       & 0.001  & 128 & 3 & 0.5 & -  \\
                Pokec            & 0.01 & 64 & 3 & 0.0 & 500000   \\
                ogbn-products    & 0.01  & 256 & 3 & 0.5 & 8192  \\
                ogbn-papers100M  & 0.001 & 256 & 3 & 0.3 & 8192 \\
                \midrule
                ogbl-ddi         & 0.005  & 256 & 2 & 0.5 & 65536   \\
                ogbl-collab      & 0.001  & 256 & 3 & 0.4 & 65536    \\
                ogbl-ppa         & 0.01 & 256 & 3 & 0.0 & 65536  \\
                \midrule
                molbbbp         & 0.001  & 300 & 5 & 0.5 & 32    \\
                moltox21        & 0.001 & 300 & 5 & 0.5 & 1000    \\
                moltoxcast      & 0.001 & 300 & 5 & 0.5 & 32    \\
                molhiv          & 0.001  & 300 & 5 & 0.5 & 32     \\
                \midrule
                molfreesolv     & 0.001  & 300 & 5 & 0.5 & 32   \\
                molesol         & 0.001 & 300 & 5 & 0.5 & 32    \\
                \bottomrule
            \end{tabular}
        \caption{Model parameters used for each dataset.}
        \label{tab:model_paramters}
        \end{table}

\section{SAGMM Architecture Components}

As illustrated in Figure~1 (main paper), \textbf{SAGMM} comprises five core components: the \textit{Node-Context Module}, \textit{Gating Mechanism}, \textit{Expert Pool}, \textit{Pruning Mechanism}, and \textit{Output Layer}. Each component is described below:

\begin{itemize}
    \item \textbf{Node-Context Module:}  
    This module processes the input node feature matrix $X \in \mathbb{R}^{N \times D}$, where $N$ denotes the number of nodes and $D$ is the feature dimensionality. To enhance the input with global contextual information, the module incorporates multihop and Laplacian-based features for node classification and link prediction tasks, and Graph Mean with Laplacian features for graph-level tasks (e.g., classification and regression). The raw node features are also directly forwarded to the expert pool as part of the forward pass.

    \item \textbf{Gating Mechanism:}  
    The gating network dynamically routes each node to a subset of experts using a thresholded attention mechanism based on SGA~\cite{wu2023sgformer}. Node representations are first transformed using learnable weight matrices and then compared via scaled dot-product similarity. The resulting attention logits are processed using a sigmoid gating function and a learnable threshold vector. A ReLU activation is then applied to retain only positive scores, thereby selecting the most relevant experts per node based on the gated activations.

    \item \textbf{Expert Pool:}  
    SAGMM leverages a diverse pool of GNN experts, each tailored to capture different structural and semantic properties of the graph. For \textit{node classification}, the expert set includes four to eight models from GCN, GAT, GraphSAGE, SGC, JKNet, GraphCNN, GIN, and MixHopNet. For \textit{link prediction}, the pool comprises GCN, GraphSAGE, JKNet, and SGC. For \textit{graph-level tasks}, the experts include GCN, GIN, GraphSAGE, and GAT. Each expert model consists of multiple layers of its respective convolution operator, enhanced with normalization (BatchNorm or LayerNorm) and residual connections where appropriate. The final node embeddings are computed as a weighted combination of outputs from the selected experts, guided by the gating mechanism.

    \item \textbf{Pruning Mechanism:}  
    To improve computational efficiency, SAGMM includes an importance-based pruning module that removes underperforming experts from the pool. This mechanism is designed to reduce both memory and inference time without degrading predictive performance.

    \item \textbf{Output Layer:}  
    The final aggregated node (or graph) embeddings are passed through a fully connected output layer to produce task-specific predictions. This layer maps the learned representations to the appropriate label space for classification or regression.
\end{itemize}

\section{Efficiency Analysis}
Table~\ref{tab:inference-time-gpu} presents the average inference time (in milliseconds) and peak GPU memory usage (in GB) during inference for various models on the ogbn-arxiv and ogbl-ddi datasets. Our proposed method, SAGMM, demonstrates a trade-off between computational cost and model complexity. While it incurs higher inference time and memory usage on ogbn-arxiv due to its more expressive gated mixture-of-experts architecture, it maintains comparable or even lower GPU memory usage on ogbl-ddi. The GPU memory footprint of SAGMM is directly influenced by the number of experts retained after pruning—more active experts result in higher memory consumption. Despite the increased overhead, SAGMM achieves significantly improved task performance (as discussed in Section 4 of main paper), justifying the additional computational cost. Interestingly, for the \textit{ogbl-collab} dataset, SAGMM achieves lower inference runtime than even the single-model baseline (GCN). This counterintuitive result is attributed to the fact that, after pruning, the remaining experts consist of lightweight architectures such as SGC and GCN with Jumping Knowledge (JK) connections. These models are either shallow or involve fewer message-passing operations. In contrast, the standard GCN baseline performs full-layer neighborhood aggregation at each layer, which can lead to higher computational overhead, particularly on dense or large graphs.

\begin{table}[ht]
  \centering
  \begin{tabular}{l|c|c|c|c}
    \toprule
    \textbf{Dataset} & \textbf{GCN (ms / GB)} & \textbf{GMoE (ms / GB)} & \textbf{DA-MoE (ms / GB)} & \textbf{SAGMM (ms / GB)} \\
    \midrule
    ogbn-arxiv     & 126 / 1.74   & 200 / 3.73   & 251 / 1.90   & 227 / 5.89 \\
    YelpChi        & 140 / 5.70   & 219 / 5.70   & 303 / 7.10   & 163 / 6.90 \\
    ogbl-ddi       & 235 / 1.50   & 298 / 1.50   & 361 / 2.15   & 334 / 1.95 \\
    ogbl-collab    & 486 / 2.54   & 631 / 6.52   & 699 / 5.78   & 464 / 5.30 \\
    ogbl-ppa       & 8555 / 6.36  & 9161 / 17.07 & 10416 / 13.08 & 9281 / 14.48 \\
    \bottomrule
  \end{tabular}
  \caption{Inference time (in milliseconds (ms)) and GPU memory usage (GB) for each model across six datasets of node classification and link prediction.}
  \label{tab:inference-time-gpu}
\end{table}

\section{Additional Experiments}


\subsection{Pre-trained Model as Experts}
\begin{figure*}[htbp]
  \centering
  \includegraphics[width=0.95\textwidth]{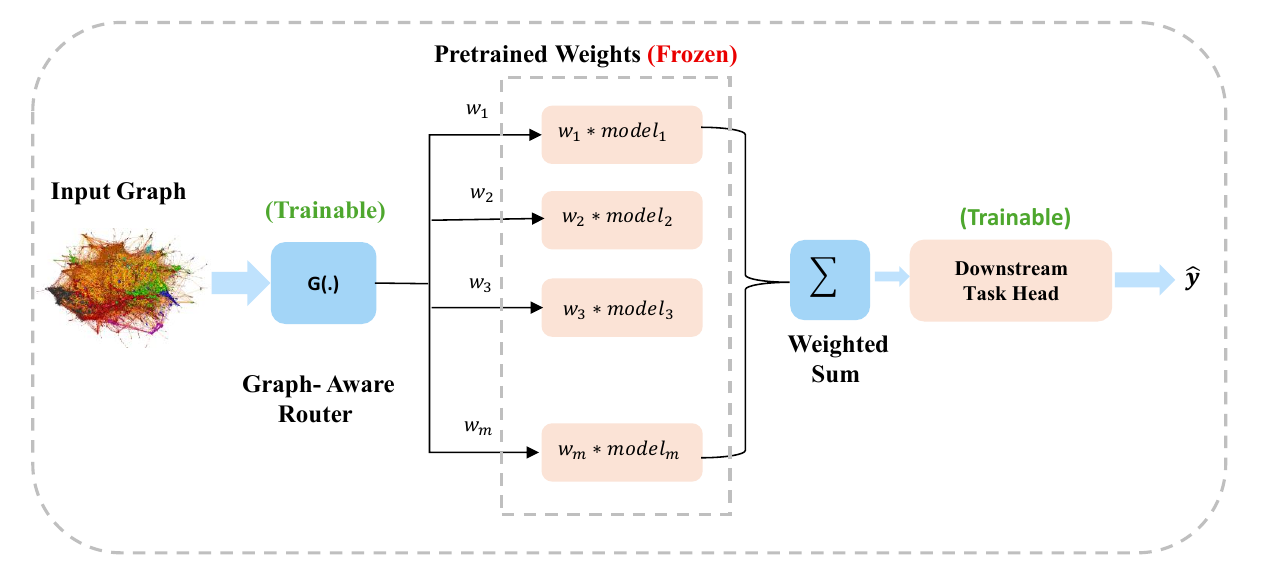} 
  \caption{SAGMM-PE architecture illustrating the modular design with pretrained GNN experts, a graph-aware gating mechanism, and a downstream task-specific head.}
  \label{fig:pretrained_model}
\end{figure*}

As shown in Figure~\ref{fig:pretrained_model}, the architecture of SAGMM-PE consists of three main components: a graph-aware router, a pool of pretrained GNN experts (Pretrained Weights), and a downstream task output head. The input graph is first passed through the router, which assigns importance scores to each expert based on the structural and feature context of the graph. These scores determine which pretrained experts to activate for a given input. Each expert has been trained independently with different pretraining ratios as listed in Table \ref{tab:results-pretrain-router}. The selected expert outputs are then combined, typically using a weighted sum based on the router’s scores, and the resulting representation is fed into an output layer that generates predictions for the downstream task. This modular design allows SAGMM-PE to efficiently reuse strong pretrained components while only training the router and output layer for the final task.
\\
We conducted pretraining using two settings: 50\% of the data and the full dataset. Importantly, the original data splits for training, validation, and testing were preserved throughout, and all evaluations were consistently performed on the full test set. In addition to varying the pretraining ratio, we also experimented with different proportions of data used for training the router. This was done to assess how much data is sufficient for the router to achieve performance competitive with or surpassing the end-to-end variant of SAGMM. As shown in Table~\ref{tab:results-pretrain-router}, our model achieves competitive results even with just 50\% of the data for pretraining and approximately 50–70\% for router training, demonstrating the data efficiency of the SAGMM framework.
\begin{table}[!htbp]
  \centering
  
  \begin{tabular}{c|c|ccccc}
    \toprule
    \textbf{Pretraining Split (\%)} & \textbf{Router Split (\%)} & \textbf{Deezer} & \textbf{YelpChi} & \textbf{ogbn-proteins} & \textbf{ogbn-arxiv} & \textbf{Pokec} \\
    \midrule
    \multirow{6}{*}{100} 
    & 10  & $63.95_{\pm 1.43}$ & $89.71_{\pm 1.46}$ & $74.60_{\pm 1.61}$ & $60.86_{\pm 0.92}$ & $67.84_{\pm 0.37}$ \\
    & 30  & $64.54_{\pm 1.46}$ & $90.35_{\pm 0.10}$ & $74.49_{\pm 1.33}$ & $66.36_{\pm 0.28}$ & $75.26_{\pm 0.29}$ \\
    & 50  & $65.23_{\pm 0.53}$ & $90.64_{\pm 0.26}$ & $75.62_{\pm 1.24}$ & $68.92_{\pm 0.35}$ & $78.41_{\pm 0.30}$ \\
    & 70  & $65.56_{\pm 0.52}$ & $91.07_{\pm 0.31}$ & $75.70_{\pm 1.34}$ & $70.42_{\pm 0.32}$ & $79.43_{\pm 0.32}$ \\
    & 90  & $65.54_{\pm 0.62}$ & $91.38_{\pm 0.19}$ & $76.31_{\pm 1.31}$ & $71.57_{\pm 0.38}$ & $81.07_{\pm 0.55}$ \\
    & 100 & $65.51_{\pm 0.57}$ & $91.33_{\pm 0.26}$ & $78.15_{\pm 1.38}$ & $72.08_{\pm 0.30}$ & $80.57_{\pm 0.95}$ \\
    \midrule
    \multirow{6}{*}{50} 
    & 10  & $63.66_{\pm 1.39}$ & $89.12_{\pm 0.56}$ &  $74.97_{\pm 2.09}$               & $61.11_{\pm 0.73}$ & $66.73_{\pm 0.32}$ \\
    & 30  & $63.62_{\pm 0.96}$ & $89.27_{\pm 0.16}$ & $74.54_{\pm 1.51}$ & $66.01_{\pm 0.49}$ & $74.24_{\pm 0.38}$ \\
    & 50  & $64.25_{\pm 0.75}$ & $89.63_{\pm 0.70}$ & $76.02_{\pm 1.16}$ & $68.28_{\pm 0.37}$ & $76.98_{\pm 0.63}$ \\
    & 70  & $64.27_{\pm 0.40}$ & $89.70_{\pm 0.67}$ & $76.11_{\pm 1.19}$ & $69.66_{\pm 0.39}$ & $77.69_{\pm 1.45}$ \\
    & 90  & $64.43_{\pm 0.73}$ & $89.92_{\pm 0.79}$ &  $76.77_{\pm 1.13}$               & $70.67_{\pm 0.31}$ & $79.60_{\pm 1.02}$ \\
    & 100 & $64.49_{\pm 0.52}$ & $89.94_{\pm 0.79}$ & $76.77_{\pm 1.07}$ & $71.02_{\pm 0.31}$ & $78.68_{\pm 1.80}$ \\
    \bottomrule
  \end{tabular}
  \caption{Performance across different pretraining and router splits on five datasets.}
  \label{tab:results-pretrain-router}
\end{table}

\section{Importance of different experts}
The use of diverse GNN experts in our architecture is motivated by their unique strengths in capturing distinct graph properties, enabling a comprehensive and adaptive representation learning process. Each expert contributes specialized capabilities that, when combined, enhance the model's ability to tackle a wide range of graph structures and tasks. Below, we summarize the key advantages of our chosen GNN models:

\begin{itemize}
    \item \textbf{GCN (Graph Convolutional Network):} GCN leverages spectral graph convolutions based on the graph Laplacian, effectively acting as a low-pass filter that smooths node features across local neighborhoods. This promotes feature homogeneity within densely connected subgraphs, making it well-suited for tasks where local connectivity patterns dominate.


    \item \textbf{GraphSAGE:} GraphSAGE employs an inductive framework that generates node embeddings by sampling and aggregating features from neighbors. This design enables efficient representation learning for unseen nodes, making it ideal for large-scale graphs. 

    \item \textbf{SGC:} Reduces complexity by removing nonlinearities and collapsing weights across layers, making it efficient for scalable learning.
    
    \item \textbf{JKNet:} GCN with jumping knowledge aggregates information from all layers, enabling the model to adaptively capture both shallow and deep node features.
    
    \item \textbf{Graph CNN:} Utilizes chebyshev polynomials to approximate the spectral convolution, allowing the model to consider larger receptive fields efficiently.
    
    \item \textbf{GIN :} Graph isomorphism network focuses on distinguishing graph structures with injective aggregation functions, making it powerful for tasks requiring structural information.
    
    \item \textbf{MixHop:} Captures multi-scale information by mixing different powers of adjacency matrices, providing enriched node representations.
    
\end{itemize}

By combining these experts, the gating mechanism adaptively assigns weights to each model's output, enabling the architecture to dynamically leverage their complementary strengths. This design ensures robust performance across diverse graph structures and tasks, balancing local feature smoothness, attention-based focus, inductive scalability, and multi-scale feature extraction.